\documentclass[journal]{IEEEtran}
\usepackage{hyperref}
\usepackage[utf8]{inputenc}
\usepackage{amssymb}
\usepackage[cmex10]{amsmath}
\usepackage{tikz}
\usepackage[export]{adjustbox}
\usepackage{graphicx}
\usepackage{xcolor}
\usepackage{multirow}
\usepackage{fancyhdr}
\usepackage{framed}
\usepackage[caption=false,font=footnotesize]{subfig}

\usepackage[linesnumbered,ruled,vlined]{algorithm2e}
\usepackage{fixltx2e}
\usepackage{comment}
\usepackage{siunitx}

\makeatletter
\newcommand*\titleheader[1]{\gdef\@titleheader{#1}}
\AtBeginDocument{%
  \let\st@red@title\@title
  \def\@title{%
    \bgroup\normalfont\large\centering\@titleheader\par\egroup
    \vskip1.5em\st@red@title}
}
\makeatother

\renewcommand{\Re}{\mathbb{R}}

\newcommand{\SL}{{\cal L}}

\newcommand{\ST}{{\cal T}}

\newtheorem{problem}{Problem}

\newcommand{\disable}[1]{}

\newcommand{\revise}[1]{{\color{blue}#1}}

\title{WiSM: Windowing Surrogate Model for Evaluation of Curvature-Constrained Tours with Dubins vehicle}

\titleheader{Under review as a paper in IEEE Transactions on Cybernetics}

\author{
   Jan Drchal, Jan Faigl, and Petr V{\'a}{\v n}a
   \thanks{The presented work has been supported by the Czech Science Foundation (GA{\v C}R) under research project No. 16-24206S. 
Access to computing at National Grid Infrastructure MetaCentrum, provided under the programme CESNET LM2015042, is greatly appreciated.
   }
   \thanks{Authors are with the Czech Technical University, 
   Faculty of Electrical Engineering, 
   Technicka 2, 166 27, Prague, 
   Czech Republic
   {\tt\small \{drchajan|faiglj|vanapet1\}@fel.cvut.cz}}%
}

\begin{document}

\maketitle

\begin{abstract}
%
Dubins tours represent a solution of the Dubins Traveling Salesman Problem (DTSP) that is a variant of the optimization routing problem to determine a curvature-constrained shortest path to visit a set of locations such that the path is feasible for Dubins vehicle, which moves only forward and has a limited turning radius.
The DTSP combines the NP-hard combinatorial optimization to determine the optimal sequence of visits to the locations, as in the regular TSP, with the continuous optimization of the heading angles at the locations, where the optimal heading values depend on the sequence of visits and vice versa.
We address the computationally challenging DTSP by fast evaluation of the sequence of visits by the proposed Windowing Surrogate Model (WiSM) which estimates the length of the optimal Dubins path connecting a sequence of locations in a Dubins tour.
The estimation is sped up by a regression model trained using close to optimum solutions of small Dubins tours that are generalized for large-scale instances of the addressed DTSP utilizing the sliding window technique and a cache for already computed results.
The reported results support that the proposed WiSM enables a fast convergence of a relatively simple evolutionary algorithm to high-quality solutions of the DTSP. 
We show that with an increasing number of locations, our algorithm scales significantly better than other state-of-the-art DTSP solvers.

\end{abstract}



\section{Introduction}
\IEEEPARstart{I}{n} this paper, we address the \emph{Dubins Traveling Salesman Problem} (DTSP) by the proposed regression surrogate model for fast evaluation of possible solution candidates given by a sequence of visits to the locations of interest by speeding up the convergence of an evolutionary solver that supports finding high-quality solutions.
The addressed DTSP is motivated by data collection and surveillance missions where a vehicle is requested to visit a set of locations as quickly as possible~\cite{oberlin2010today,faigl19jfr}.
The problem is a variant of the routing problem~\cite{niendorf_exact_2018} to determine the optimal sequence of visits to the given set of locations such that the length of the curvature-constrained path connecting the locations in the sequence is minimal~\cite{savla2005point}.
The curvature-constrained path is requested because of motion constraints of the vehicle that are modeled by Dubins vehicle~\cite{dubins1957curves} which moves only forward with the limited turning radius~$\rho$.
A solution of the DTSP consists of sequencing part to determine the optimal sequence of visits to the locations and continuous optimization to determine the optimal heading values.
The DTSP is at least NP-hard~\cite{ny12} because for $\rho=0$, it becomes the \emph{Euclidean TSP} (ETSP)~\cite{applegate07}.

In 1957, Dubins showed that for two given locations with the prescribed leaving and arrival heading angles of the vehicle, the optimal curvature-constrained path connecting the locations is one of six possible maneuvers that can be found by a closed-form solution~\cite{dubins1957curves}. 
The generalization to an arbitrary number of locations is known as the \textit{Dubins Touring Problem} (DTP).
The state-of-the-art DTP solvers are based on a sampling of the possible heading angles~\cite{manyam2017tightly}.
In~\cite{faigl17ecmr}, an informed sampling method, which we further denote the \textit{Iteratively-Refined Informed Sampling} (IRIS), enables finding a solution of the DTP close to optimum, i.e., with the relative optimality gap to the lower bound less than 1\%, in tens of seconds for problems with up to 100 locations, which outperforms the uniform sampling~\cite{manyam2017tightly}.

The DTSP solvers reported in the literature can be roughly classified into three groups according to~\cite{macharet18survey,vana15iros}.
The first group represents \textit{decoupled} approaches, in which the sequence of visits is determined independently from the headings, e.g., using a solution of the related ETSP.
Then, headings are determined in the second step using a heuristic Alternating Algorithm (AA)~\cite{savla2005point}, the Local Iterative Optimization (LIO)~\cite{vana15iros}, or by the IRIS proposed in~\cite{faigl17ecmr}.
The methods of the second group involve a \textit{joint} optimization of both the sequence and the headings using Evolutionary Algorithm (EA) as it is reported in~\cite{macharet12evolutionary,zhang2014memetic}. 
The last group includes approaches based on \textit{transformation} of the original problem into a purely combinatorial routing problem.
In~\cite{oberlin2010today} and~\cite{obermeyer2012sampling}, the heading angles are sampled, and the discretized instance is transformed to the Generalized TSP that is further transformed into the Asymmetric TSP that can be solved by the heuristic LKH solver~\cite{lkh}, or Concorde solver~\cite{concorde} for the optimal solution with respect to the selected sampling. 
Similar sampling-based strategy is utilized to find a tight lower bound of the DTSP in~\cite{manyam2018tightly}.

The existing \textit{decoupled} approaches such as~\cite{manyam2017tightly,faigl17ecmr} use only a single sequence found as a solution of the Euclidean TSP without considering the minimum turning radius $\rho$ of the Dubins vehicle which limits the quality of the DTSP solutions for cases where $\rho$ cannot be ignored (i.e., dense location distribution w.r.t. $\rho$).
The \textit{joint optimization} and \textit{transformation} methods may provide better solutions at the cost of low scalability (dense sampling leads to large problem instances of the transformed problems) and very high computational requirements because a large search-space needs to be covered by the global optimizer such as the evolutionary algorithm.

The herein presented approach is a novel DTSP solver that combines features of both the \textit{decoupled} approaches and \textit{joint optimization}.
The main idea of the proposed method is that a relatively simple EA optimizes the sequence of visits separately from the continuous optimization part, and the sequence quality (i.e., fitness) of each individual is assessed by means of length of the corresponding DTP solution (which also determines the heading angles).
The information about headings is not stored in the genome itself, but rather computed during every evaluation of its fitness.
It reduces the complexity of the search-space and makes it strictly combinatorial as the sequence of visits is a permutation; however, it also makes the evaluation of each individual more expensive.
Thus, the critical part of our method is the need to evaluate possibly many candidate sequences.
Even though the solution of the DTP by the IRIS~\cite{faigl17ecmr} is significantly faster than the uniform sampling~\cite{manyam2017tightly}; it is still too computationally demanding.

Similarly to~\cite{wang_global_2018}, where the authors speed up the expensive evaluation of the objective function using a surrogate model, we propose the \textit{Windowing Surrogate Model} (WiSM) leveraging on the high-quality solutions of the IRIS-based DTP solver.
The main contribution of the proposed WiSM is considered in the windowing decomposition developed from a lower bound which allows training of the model on small fixed-size DTP instances, while the evaluation of the solution length can be performed on DTP instances of arbitrary size.
It makes WiSM easily implementable by most types of regression models. 

Based on the presented results, WiSM provides solutions competitive to a very dense sampling of the heading values in the DTP, but with significantly lower computational requirements.
The paper is mainly focused on the surrogate model based on the windowing decomposition and not on details and tuning of the sequence optimization; however, we show that the combination of WiSM with a simple EA (further denoted as WiSM-EA) outperforms other DTSP solvers for a wide range of instance sizes.

The paper is organized as follows.
The addressed DTSP is defined in the next section.
A brief overview of the IRIS, the high-quality DTP solver used in learning of the proposed WiSM, is presented in Section~\ref{sec:dtp}.
The model is employed in the Evolutionary Algorithm summarized in Section~\ref{sec:method_ea}.
The surrogate model itself with the windowing decomposition is proposed in Section~\ref{sec:method}.
Empirical evaluation of the WiSM-EA and its comparison with existing approaches for the DTSP are in Section~\ref{sec:results}.
Concluding remarks are given in Section~\ref{sec:conclusion}.


\section{Problem Statement\label{sec:problem}}
The proposed solution of the DTSP is based on an evaluation of the candidate sequences to visit the given set of $n$ target locations $\mathcal{T}=\{t_1,\ldots,t_n\}$, $t_i\in\mathbb{R}^2$ using a solution of the DTP where the path connecting the locations $\mathcal{T}$ has to respect the motion constraints of Dubins vehicle~\cite{dubins1957curves} that is moving only forward with a constant forward velocity $v$ and a minimum turning radius $\rho$. 
The state of the vehicle can be expressed as $q=(x, y, \theta)$, where $(x, y) \in \Re^2$ is the position of the vehicle and $\theta \in \mathbb{S}^1$ is its heading angle. Thus the state $q$ is from the Special Euclidean group, $q\in SE(2)$.
The motion of the vehicle can be described as
\begin{equation}
   \label{eq_dubins_vehicle}
   \dot{q} 
   =
   \begin{bmatrix}
      \dot{x}\\
      \dot{y}\\
      \dot{\theta}\\
   \end{bmatrix} 
   = 
   v
   \begin{bmatrix}
      \cos \theta\\
      \sin \theta\\
      \frac{u}{\rho}\\
   \end{bmatrix},\;
   \quad  
   \vert u\rvert \leq 1,
\end{equation}
where $u$ is the bounded control input. 
For simplicity and w.l.o.g., we further consider $v = 1$ and $\rho = 1$.

A solution of the DTSP can be expressed as a permutation $\Sigma=(\sigma_1,\ldots, \sigma_n)$ for $1 \le \sigma_i \le n$ that defines a sequence of visits to the locations $\mathcal{T}$ and particular heading values $\Theta = \{\theta_1,\ldots,\theta_n\}$ corresponding to each particular location.

The DTSP stands to find a sequence of visits $\Sigma$ to $\mathcal{T}$ with the respective heading values $\Theta$ such that the length $C(\Sigma, \Theta)$ of the Dubins path connecting the locations $\mathcal{T}$ is minimal.
The problem can be considered as combinatorial optimization over all possible sequences $\Sigma$ and $n$-variable continuous optimization of the heading values $\Theta$.

\vspace{0.5em}
\begin{problem}[Dubins Traveling Salesman Problem - DTSP]
   \label{prob_open_DTSP}
   \begin{eqnarray}
      \label{eq:open}
      \operatorname{minimize}_{\,\Sigma,\Theta} 
      &C(\Sigma, \Theta) = \displaystyle\sum_{i=1}^{n}\SL(q_{\sigma_i}, q_{\sigma_{i+1}})\\
      \operatorname{s.t.}\nonumber\\
      &\begin{array}{l}
     q_i=(t_i,\theta_i),\, q_i\in SE(2),\, t_i\in \mathcal{T},\\ 
     \Sigma=(\sigma_1,\ldots,\sigma_n),\, \sigma_i \in \{1,\ldots, n\}, \\
     \sigma_i\neq\sigma_j\text{ for } i\neq j, \\ 
     \Theta=\{\theta_1,\ldots, \theta_n\}, 
      \end{array}\nonumber
   \end{eqnarray}
   where $\SL(q_i, q_j)$ denotes the analytically computed length of the optimal Dubins maneuver~\cite{dubins1957curves} between $q_i$ and $q_j$, and $\sigma_{i} \triangleq \sigma_{i-n}$ for $i > n$ is defined to simplify the notation in (\ref{eq:open}).
\end{problem}

Finding a solution of the DTSP is an optimization problem with a combinatorial part over $\Sigma$ and continuous part over $\Theta$.
If $\Sigma$ is given, the problem becomes strictly continuous optimization with $n$ variables in $\Theta$. 
The problem is then called the \textit{Dubins Touring Problem} (DTP), and the solution cost for a particular sequence $\Sigma$ is denoted $C(\Sigma)$. 
\vspace{0.5em}
\begin{problem}[Dubins Touring Problem - DTP]
   \begin{eqnarray}
      \label{eq:dtp}
      C(\Sigma) = \min_{\Theta} C(\Sigma, \Theta).
   \end{eqnarray}
\end{problem}

Due to the combinatorial nature of the TSP, many candidate sequences need to be evaluated.
Thus, the evaluation of (\ref{eq:dtp}) should be fast enough to provide competitive results to existing DTSP approaches~\cite{macharet18survey}.
In Section~\ref{sec:method}, we propose a surrogate approximator of (\ref{eq:dtp}) trained using high-quality solutions of the DTP found by the IRIS method~\cite{faigl17ecmr} briefly described in the following section.


\section{Background -- Dubins Touring Problem (DTP)\label{sec:dtp}}

A solution of the DTP is needed to evaluate the cost (\ref{eq:dtp}), and thus the final solution of the DTSP, but the solution is also needed to train the proposed surrogate model for fast evaluation of possible candidate sequences. 
Finding optimal heading angles for the DTP with dense locations is a challenging problem because the length of the Dubins tour connecting a sequence of locations is a piecewise continuous function~\cite{goaoc2013bounded}.
Sampling-based approaches are thus utilized to sample the domains of the continuous variables into finite sets of discrete values of possible heading angles.
A solution can be then found as the shortest path in a graph representing the discretized instance of the DTP as follows.

Let have $k$ samples of the heading angle per each location $t_i\in\mathcal{T}$ which form $k$ possible states per each $t_i$.
Then each pair of the states corresponding to two consecutive locations in the sequence can be connected by the optimal curvature-constrained path determined as the optimal Dubins maneuver~\cite{dubins1957curves}.
Having the states and their connections, a search graph can be created, and the shortest path for the given sequence of $n$ locations and defined sampling $k$ can be found in $O(nk^3)$ by a forward search based on dynamic programming~\cite{faigl17ecmr}.
Such a path is a feasible solution to the DTP.

High-quality solutions, however, require dense sampling~\cite{manyam2017tightly}, which can be computationally demanding.
The IRIS method~\cite{faigl17ecmr} decreases the computational requirements by an iterative refinement of the heading intervals based on the tight lower bound solution from~\cite{manyam2017tightly}.

The DTSP solver presented in this paper requires a high number of sequence evaluations, which makes even the IRIS approach~\cite{faigl17ecmr} too computationally demanding.
Therefore, we propose to employ a surrogate model for fast evaluation of Dubins tour lengths to speed up the convergence towards high-quality DTSP solutions using generated sequences by the Evolutionary Algorithm described in the following section.


\SetKw{not}{not}
\SetKwRepeat{Do}{do}{while}
\SetKwFunction{initializePopulation}{initializePopulation}
\SetKwFunction{initializeArchive}{initializeArchive}
\SetKwFunction{evaluatePopulation}{evaluatePopulation}
\SetKwFunction{updateArchive}{updateArchive}
\SetKwFunction{stopTime}{$T$\textsubscript{S}}
\SetKwFunction{Tcpu}{\text{$T$\textsubscript{CPU}}}
\SetKwFunction{stopCondition}{stopCondition(\stopTime)}
\SetKwFunction{best}{best}
\SetKwFunction{sort}{sort}
\SetKwFunction{mutation}{mutation}
\SetKwFunction{crossover}{crossover}
\SetKwFunction{tournamentSelection}{tournamentSelection}
\SetKwFunction{DTPCost}{DTPCost}
\SetKwFunction{DTPSolution}{DTPSolution}
\SetKwFunction{random}{random}

\section{Evolutionary Algorithm for the DTSP}\label{sec:method_ea}

In this section, we present the Evolutionary Algorithm (EA) for solving the DTSP.
Contrary to existing evolutionary approaches to the DTSP, such as \cite{zhang2014memetic} where sequences and headings are encoded together, the proposed evolutionary approach is utilized only for generating proper sequences $\Sigma$, and the heading values are determined as a solution of the corresponding DTP.
The particular cost function $C(\Sigma)$ can be determined by IRIS for the DTP~\cite{faigl17ecmr}, but also by its surrogate model estimation proposed in Section~\ref{sec:method}.
A~generational EA with a simple elitism, where an individual of the population is represented by a permutation of $n$ target locations (path representation~\cite{larranaga1999}) is used, and it is summarized in Algorithm~\ref{alg:ea}.

\begin{algorithm}[!htb]
   \caption{Evolutionary Algorithm for the DTSP}
   \label{alg:ea}
   \SetKwInOut{Input}{Input}\SetKwInOut{Output}{Output}
   \Input{$\ST$ -- a set of the given $n$ target locations;
   $N$ -- population size; 
   $t$ -- tournament size;
   $p_m$ -- mutation probability;
   $e$ -- elite size;
   $\stopTime$ -- termination time;
   }
   \Output{DTSP solution -- ($\Sigma$, $\Theta$, $C(\Sigma)$)}
   \BlankLine
   $P \leftarrow$ \initializePopulation{$N$, $n$}\label{alg:ea_init}\;
   \evaluatePopulation{$P$, $\ST$}\;
   \While{\Tcpu$<\stopTime$}{\label{alg:ea_loop}
   $P_{new} \leftarrow \{\}$\label{alg:ea_new}\;
   \For{$i\leftarrow 1$ \KwTo $N$}{
      $a_1 \leftarrow$\tournamentSelection{$P$, $t$}\;
      \eIf{\random{} $< p_m$}{
     $c \leftarrow$ \mutation{$a_1$}\;
      }{
     $a_2 \leftarrow$\tournamentSelection{$P$, $t$}\;
     $c \leftarrow$ \crossover{$a_1$, $a_2$}\;
      }
      $P_{new} \leftarrow P_{new}\cup\{c\}$\;
      \label{alg:ea_new_end}
   }
   \evaluatePopulation{$P_{new}$, $\ST$}\label{alg:ea_eval}\;
   $b \leftarrow $\best{$P \cup P_{new}$}\;
   $P \leftarrow \{\}$\;
   \lFor{$i\leftarrow 1$ \KwTo $e$}{\label{alg:ea_elite}
   $P \leftarrow P \cup \{b\}$
   }
   $B \leftarrow $\sort{$P_{new}$}\label{alg:ea_sort}\;
   \lFor{$i\leftarrow 1$ \KwTo $N-e$}{\label{alg:ea_rest}
   $P \leftarrow P \cup \{B[i]\}$
   }
   }
   $\Sigma \leftarrow b$\;\label{alg:ea_seq}
   $(\Theta, C(\Sigma)) \leftarrow \text{DTP}(\Sigma)$\;\label{alg:ea_dtp}
   \Return $(\Sigma, \Theta, C(\Sigma))$\;
\end{algorithm}

The initial population is filled by random permutations in the \initializePopulation function representing sequences of visits to the given locations.
The fitness of each individual in the population is evaluated using the solution (or its estimation) of the corresponding DTP by the \evaluatePopulation function.
The main loop (Line~\ref{alg:ea_loop}, Algorithm~\ref{alg:ea}) iterates over the generations until the overall running time \Tcpu reaches the termination time~$\stopTime$.

The evolution is performed as follows.
A new population of $N$ individuals is generated either by \mutation with the probability of $p_m$, or \crossover otherwise (Lines~\ref{alg:ea_new}--\ref{alg:ea_new_end}, Algorithm~\ref{alg:ea}).
The \mutation operator is the \emph{Simple Inversion Mutation} (SIM)~\cite{holland1992,larranaga1999} which reverses a random sub-sequence.
The \crossover method implements the well-known Order crossover (OX)~\cite{davis1985} which copies a random sub-sequence of the first parent and then adds locations from the other parent preserving their order but leaving out those already used.
We specifically use the OX1 variety as described in~\cite{larranaga1999}, although we generate only a single offspring.
Both \mutation and \crossover employ tournament selection with the tournament size $t$ (\tournamentSelection) to select the parent(s).

The new population is then evaluated (Line~\ref{alg:ea_eval}, Algorithm~\ref{alg:ea}) and the best solution of the both current $P$ and new $P_{new}$ populations is selected as $b$. 
Finally, $e$ least viable individuals of $P_{new}$ are replaced by the copies of the best so far solution $b$ introducing the elitism (Lines~\ref{alg:ea_elite}--\ref{alg:ea_rest}, Algorithm~\ref{alg:ea}).

The algorithm terminates when the stop condition is met. 
The best sequence $\Sigma$ is extracted from the population (Line~\ref{alg:ea_seq}, Algorithm~\ref{alg:ea}) and the corresponding headings $\Theta$ are determined (Line~\ref{alg:ea_dtp}, Algorithm~\ref{alg:ea}) calling the DTP solver~\cite{faigl17ecmr} to provide a feasible solution.
On the other hand, sampling-based approaches would be too computationally demanding for evaluation of $C(\Sigma)$ of all population individuals in \evaluatePopulation, and therefore, we propose a surrogate model to get a considerable speedup.



\section{Proposed Windowing Surrogate Model (WiSM) for a Fast Estimation of the DTP Solution Cost\label{sec:method}}

The proposed \textit{Windowing Surrogate Model} (WiSM) is designed to approximate $C(\Sigma)$ with a surrogate function to significantly speed up the evaluation of the DTP instance costs.
The WiSM is evaluated based on overlapping windows in a convolutional manner, and the partial results are averaged to get the final cost estimation.

In the studied DTSP, the number of locations $n$ can vary depending on the particular problem instance.
Thus, we need to address the limitation of the regression models (e.g., neural networks, random forests, etc.) that are mostly limited to fixed sized inputs.
While this issue can be directly approached by recurrent neural networks that allow processing inputs of variable-length as sequences (e.g., LSTM~\cite{hochreiter1997_lstm} or GRU~\cite{cho-etal-2014-learning}), a collection of training data would be non-trivial as we would need to collect examples of DTP instances having a varying number of locations with a little guarantee on how the network would generalize on sequences of the unseen lengths.
To overcome the problem of the variable-size input, we propose a decomposition of the cost function $C(\Sigma)$ to fixed-size subproblems based on the sliding window technique.

\subsection{Sliding Windowing based Cost Estimation of the DTP\label{sec:method_cost_decompositon}}
The proposed idea is to evaluate the cost of Dubins tour for small sub-sequences of the target locations limited by the window size $w$.
The total tour cost can be then estimated as an aggregate of the particular costs.

For a specific sequence $\Sigma$ we can define the particular cost $C^*_{w,i}(\Sigma)$ of the subtour of Dubins optimal tour for the fixed $w$-size window as 
\begin{equation}\label{eq:optc}
   C^{*}_{w,i}(\Sigma)= \sum_{j=i}^{i+w-1}\SL(q^*_{\sigma_j}, q^*_{\sigma_{j+1}}),\;\; i\in \{1, \ldots, n\}.
\end{equation}

\begin{figure}[!htb]
   \centering
   \includegraphics[width=0.99\columnwidth,trim={1.9cm 0 1.9cm 0},clip]{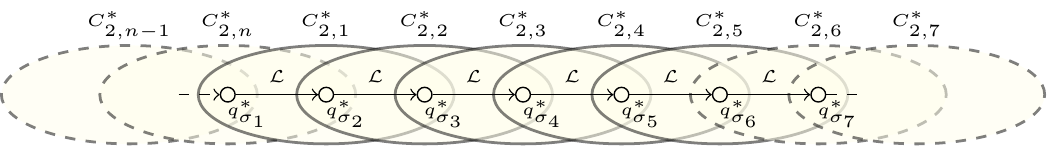}
   \caption{Sliding window principle with overlapping windows of the size $w=2$. 
   Each window covers 2 trips, i.e., 3 locations.\label{fig:sliding_window}
   }
\end{figure}

We propose to utilize a sliding window that starts at each target location $i$ as it enables to address arbitrarily-length instances for which the number of locations $n$ is not necessarily divisible by the utilized window size $w$.
The whole Dubins tour is divided into $n$ sub-sequences of the windows $(\sigma_{i}, \ldots, \sigma_{i+w})$ for $i\in\{1,\ldots,n\}$ each connecting $w+1$ successive locations as it is depicted in Fig.~\ref{fig:sliding_window}.
Then the optimal cost $C^*(\Sigma)$ for the specific $\Sigma$ and known optimal heading angles $\Theta^*$ can be computed from the window costs exactly as
\begin{equation}\label{eq:exact}
   C^*(\Sigma) = \frac{1}{w}\sum_{i=1}^{n}C^*_{w,i}(\Sigma).
\end{equation}
The optimal heading angles $\Theta^*$ are apparently not available, and thus (\ref{eq:exact}) provides only an intuition behind the proposed approach for fast cost estimation of the Dubins tour.

The cost of the specific window is estimated independently on the final solution and finding the cost of the window sub-sequence is considered as the \textit{Open DTP}~\cite{faigl17ecmr}, which can be defined as the continuous optimization problem
\begin{equation}
   \label{eq:dtp_open}
   \overline{C}(\Sigma) = \min_{\Theta} \overline{C}(\Sigma, \Theta).
\end{equation}
The Open DTP is further denoted as $\overline{\text{DTP}}$ to emphasize it does not involve a closed tour and the heading angles at both end locations are unconstrained.
Note that the unlike in (\ref{eq:open}), the length of the closing Dubins maneuver $\SL(q_{\sigma_n}, q_{\sigma_{1}})$ is omitted from the $\overline{C}(\Sigma, \Theta)$ definition:
\begin{equation}
   \overline{C}(\Sigma, \Theta) = \displaystyle\sum_{i=1}^{n-1}\SL(q_{\sigma_i}, q_{\sigma_{i+1}}).
\end{equation}
Comparing the particular DTP cost $C^*_{w,i}(\Sigma)$ of the $i$-th window defined by a sub-sequence $\Sigma_{w,i} \triangleq (\sigma_i, \ldots, \sigma_{i+w})$ and the corresponding $\overline{\text{DTP}}$ cost $\overline{C}(\Sigma_{w,i})$, it can be shown that
\begin{equation}
   \label{eq:single_lower_bound}
   \overline{C}(\Sigma_{w,i}) \leq C^*_{w,i}(\Sigma).
\end{equation}
The proof by contradiction is based on the fact that $\overline{C}(\Sigma_{w,i})$ is a relaxed version of $C^*_{w,i}(\Sigma)$.
More specifically, unlike for $C^*_{w,i}(\Sigma)$, the boundary angles $\theta^*_i$ and $\theta^*_{i+w}$ of $\overline{C}(\Sigma_{w,i})$ are not constrained.
Thus $(q_{\sigma_1},q_{\sigma_2})$ and $(q_{\sigma_{n-1}},q_{\sigma_n})$ can take advantage of arbitrarily chosen heading angles at the endpoints.

\begin{figure}[!htb]
   \centering
   \includegraphics[height=4cm, valign=c]{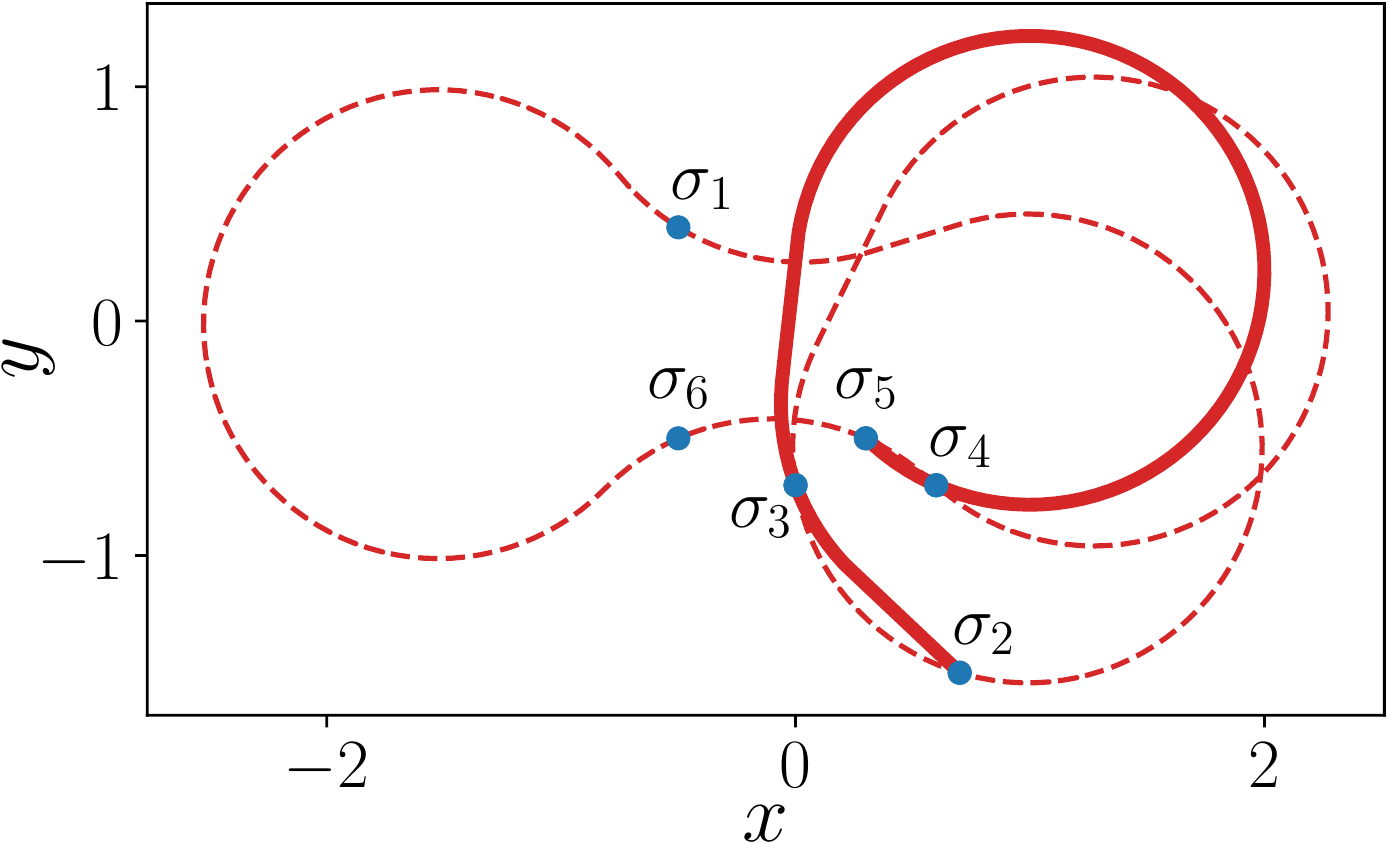}
   \caption{
      A solution of the closed DTP (dashed line) with one of $n$ open DTP subtours $\Sigma_{3,2}$ (thick solid line).
   }
   \label{fig:open_subtour_example}
\end{figure}
An example of the Closed DTP with $n=6$ locations with a corresponding instance of the Open DTP over four locations $\sigma_2, \sigma_3, \sigma_4$, and $\sigma_5$ is depicted in Fig.~\ref{fig:open_subtour_example}.
Note that the optimal tours between $\sigma_2$ and $\sigma_5$ differ for the open and closed cases. 
The open version is  shorter which corresponds with (\ref{eq:single_lower_bound}).

The cost of the whole sequence $\Sigma$ can be estimated as $\tilde{C}_w(\Sigma)$ using $n$ windows with the size $w$ as
\begin{equation}
   \label{eq:cost}
   \tilde{C}_w(\Sigma) = \frac{1}{w}\sum_{i=1}^{n}\overline{C}(\Sigma_{w,i}).    
\end{equation}
Based on (\ref{eq:single_lower_bound}), the estimation $\tilde{C}_w(\Sigma)$ is a lower bound of the optimal cost $C^*(\Sigma)$ and we get
\begin{equation}
   \label{eq:cost_lb}
   \tilde{C}_w(\Sigma) \leq C^*(\Sigma).
\end{equation}
In the following, we expect the lower bound (\ref{eq:cost}) is tight enough even for relatively small size $w$ and the minimization w.r.t. $\tilde{C}_w(\Sigma)$ can be employed as a proxy to the direct minimization of $C(\Sigma)$.
Thus, we can evaluate DTP costs for instances of a fixed size $w + 1$ and use (\ref{eq:cost}) to get the cost estimate for an instance of arbitrary size $n$.

The computational complexity of $\tilde{C}_w(\Sigma)$ approximation employing the sampling methods such as IRIS~\cite{faigl17ecmr} can be bounded by $O(nwk^2)$, where $k$ is the number of heading values per each location.
In practice, the window size $w$ is expected to be constant and significantly lower than both $k$ and $n$ ($w\ll k$ and $w\ll n$).
Thus, the computational complexity can be bounded by $O(nk^2)$, which is a noticeable improvement in comparison to $O(nk^3)$ for the evaluation of $C(\Sigma)$.
While the performance improvement is appealing, it turns out, that in practice, the sampling methods are still too slow to allow for efficient evolutionary search. 
Nevertheless, the decomposition, based on a repeated evaluation of fixed-sized sub-tours, enables approximation of the relatively slow sampling-based methods by a wide selection of regression models. 

The utilized sliding window approach is a commonly used general technique to process variable length data; however, to the best of the authors' knowledge, it has not been used for tour length estimates yet.
In the particular case of the proposed WiSM, the specific combination of the sliding window with the selected aggregation scheme gave a clear interpretation (\ref{eq:cost_lb}), which is only possible by incorporating domain-specific knowledge of the addressed DTSP using results of the DTP.

\subsection{Windowing Surrogate Model (WiSM) Approximation of the Cost Estimation $\tilde{C}_w(\Sigma)$\label{sec:method_dtp_approx}}

The key idea of the proposed WiSM-based approximation of the DTP solution cost is based on a fast evaluation of the partial costs of $\overline{C}(\Sigma_{w,i})$ by a surrogate regression model.
In existing deployments of the surrogate models in evolutionary algorithms, the model is a proxy to the \textit{real fitness} that is trained online while EA runs.
The switching between the real and surrogate fitness can be realized by more or less complex \textit{model management}~\cite{jin2005, jin2011}. 
On the other hand, the real-time response of the proposed DTSP solver can benefit from pre-trained surrogate model~\cite{horng15cyb}, and therefore, the model is learned offline using a large training dataset based on high-quality solutions of the open-loop $\overline{\text{DTP}}$.

The WiSM can be implemented using virtually any type of regression model, but we consider the Multi-Layer Perceptron (MLP)~\cite{bishop2006} in this paper.
It is because the feed-forward neural network supports fast evaluation of all partial costs $\overline{C}(\Sigma_{w,i})$ in a single batch.
Thus, it can take advantage of highly optimized matrix multiplication routines\footnote{We can evaluate the whole population of the evolutionary algorithm in a single batch, which has been utilized for the herein reported empirical results.}.

Each input vector fed to the surrogate model is constructed as $(x_{\sigma_i}, y_{\sigma_i}, \ldots, x_{\sigma_{i+w}}, y_{\sigma_{i+w}}) \in \Re^{2(w+1)}$ while the output constitutes single real value corresponding to $\overline{C}(\Sigma_{w,i})$.
Notice, the number of locations is $w+1$ because the cost of Dubins tour is defined for two locations and more, see (\ref{eq:optc}).
Hence, the MLP architecture has $2(w+1)$ inputs, several hidden layers of non-linear units such as ReLU~\cite{glorot2011_relu}, and a single linear output neuron which is common for regression tasks.


\section{Results\label{sec:results}}
The proposed Windowing Surrogate Model (WiSM) has been evaluated in combination with the evolutionary algorithm presented in Section~\ref{sec:method_ea} and the combined method is further denoted WiSM-EA.
The solved DTSP instances are of various sizes and different densities of the locations.

The performance of the WiSM-EA is compared with the existing DTSP approaches. 
Namely, the Alternating Algorithm (AA)~\cite{savla2005point}, the Local Iterative Optimization (LIO)~\cite{vana15iros}, and the Sampling-based Algorithm (SA)~\cite{obermeyer2009path}.
The SA transforms the problem to the Generalized TSP that is further transformed into Asymmetric TSP solved by the LKH solver~\cite{lkh}.

As a \textit{baseline}, we take the decoupled approach from~\cite{faigl17ecmr} where the DTSP solution takes the location sequence $\Sigma$ from the solution of the corresponding Euclidean TSP (obtained by the Concorde solver~\cite{concorde}). 
The headings $\Theta$ are consequently determined by the IRIS method.
Besides, we also initially considered the Memetic algorithm~\cite{zhang2014memetic}; however, the achieved results are not competitive with the selected approaches regarding the solution quality and computational requirements.
Therefore it is not included in the reported results.

The DTSP methods were compared using instances, randomly generated similarly to~\cite{faigl17ecmr}. 
The benchmark set consists of 10 instances per each possible pair $(n, d)$, where $n \in \{25, 100, 500\}$ denotes the number of target locations and $d\in \{0.1, 1, 10\}$ is the density of the locations.
The targets $\mathcal{T}$ of each instance were uniformly sampled from a square region with a side $b=\sqrt{n/d}$, i.e., $t_i \in\left[-\frac{1}{2}b, \frac{1}{2}b\right] \times \left[-\frac{1}{2}b, \frac{1}{2}b\right]$.

Due to unknown optimal solutions of the DTSP instances, the results are evaluated using the \textit{normalized cost} defined as 
\begin{equation}
   \label{eq:normalized_cost}
   C_r= \dfrac{C(\Sigma, \Theta)}{C(\Sigma_{\text{base}}, \Theta_{\text{base}})},
\end{equation}
where $C(\Sigma_{\text{base}}, \Theta_{\text{base}})$ is the cost provided by the baseline.

The headings $\Theta$ for the baseline and also for the WiSM-EA (Line~\ref{alg:ea_dtp}, Algorithm~\ref{alg:ea}) were determined using IRIS~\cite{faigl17ecmr} with the maximum number of samples set to $k_\text{max}=1024$.

The parameters used by the evolutionary algorithm were: the population size $N = 100$, tournament size $t = 3$, mutation probability $p_m = 0.8$, and elitist size $e = 20$.

The WiSM-EA was implemented in Julia language~\cite{julia}, run with \texttt{\mbox{-O3}} and \texttt{\mbox{--check-bounds=no}} options and the computational requirements were further significantly decreased by caching model approximations $\overline{C}(\Sigma_{w,i})$ using $\Sigma_{w,i}$ as a key.
The IRIS method (called from Line~\ref{alg:ea_dtp}, Algorithm~\ref{alg:ea}) and also all the other algorithms (AA, LIO, SA) were implemented in C++ and compiled by the gcc compiler with \texttt{\mbox{-O3}} and \texttt{\mbox{-march=native}} flags, which also holds for the LKH.
In all the cases, the algorithms were run on a single core of the Intel Xeon Gold 6130 @ \SI{2.10}{GHz} processor.

\subsection{Learning Setup of the Proposed WiSM\label{sec:experiments_setup}}
The choice of an appropriate window size $w$ of the WiSM deals is a trade-off between the model complexity, accuracy, and computational time needed to evaluate the model.
Increased window size leads to the increased number of the model inputs and parameters that also increases demands on the size of the training dataset.
Besides, a prediction for a complex model is computationally demanding which might be critical for the evolutionary search because it can limit the total number of evaluations achievable under the particular computational time limit \stopTime (Line~\ref{alg:ea_dtp}, Algorithm~\ref{alg:ea_loop}).

Empirical evaluation has been performed for $2 \le w \le 10$ in the same way as described in this section, but data are not shown for brevity.
Based on that, we found out that $w=3$ provides overall best results, and therefore, this window size is considered for the rest of the presented results.
The achieved average computational time per a single window for $w = 3$ is $\SI{0.2}{\mu s}$ and $\SI{3.3}{\mu s}$ with and without the caching, respectively.

The empirically selected neural network architecture was the MLP as described in Section~\ref{sec:method_dtp_approx} having three hidden layers of 256 ReLU units each.  
The training dataset consisted of $64\times 10^6$ samples of $2(w+1) = 8$ coordinates and the corresponding target cost value was computed using a solution of the open DTP found by the IRIS method with $k_\text{max}=1024$.
The individual coordinates of the learning datasets were independently drawn from the normal distribution ${\cal N}(0, 1)$ which introduced a small fraction of longer distances between the target locations improving generalization overall  considered densities $d$ of the benchmark instances\footnote{It would take roughly 21 days to generate the full dataset for $w = 3$ on a single core, and thus 64 cores were used to speed up the process.}. 
The distribution was also selected empirically, where the process was bootstrapped by observing the distributions of locations in sub tours of the random DTSP instances. 
We found WiSM-EA to be reasonably robust w.r.t. to the training distributions; nevertheless, future research should focus on training data generation methods.

The model training was done as follows. 
We split the dataset to training part ($80\%$) and the validation part ($20\%$).
The loss function to train the surrogate model was the Mean Squared Error (MSE), which is a common choice for the regression.
No regularization method such as L1, L2, or drop out~\cite{srivastava2014} was used as overfitting was not a problem due to the size of the training data. 
We employed Adam~\cite{kingma2015adam} using recommended parameter values (the learning rate $0.001$, $\beta_1=0.9$ and $\beta_2=0.999$). 
The training was terminated using \textit{early stopping} when there was no improvement in the validation loss for ten successive epochs. 
The single WiSM employed for all $(n, d)$ pairs in the following experiments needed 134 epochs to converge, achieving MSE of $1.94\times 10^{-2}$ on the validation set\footnote{The target values were standardized as usual when training neural networks.}.

\subsection{WiSM-EA Performance Evaluation\label{sec:main_results}}
The evaluation results of the proposed WiSM-EA and its comparison with the other DTSP solvers is depicted in Figures~\ref{fig:results_d1_n100}, \ref{fig:results_densities} and~\ref{fig:results_locations}.
Each data point represents the mean value of the real required computational time and the mean value of the corresponding \textit{normalized cost} $C_r$ computed over 100 runs of the particular algorithm, i.e., ten runs per each of the ten random instances for each problem setup $(n, d)$.
The boxes in the presented plots delimit lower and upper quartiles. 
Note that, in general,  $C_r$ increases with the decreasing density of $d$ as the problem becomes more similar to its underlying Euclidean TSP.

The WiSM and SA methods were run using multiple settings in order to study the tradeoff between the solution quality and computational requirements.
In particular, for the WiSM-EA, we used eight values of the time limit \stopTime (evaluated by the \stopCondition method) ranging from 1 second to 600 seconds, denoting the algorithm configuration as WiSM-EA\textsubscript{\stopTime}.
The total required computational time $\Tcpu > \stopTime$, as in its final stage WiSM-EA calls IRIS to compute the headings $\Theta$ for the sequence $\Sigma$ found by the evolutionary search (Line~\ref{alg:ea_dtp}, Algorithm~\ref{alg:ea}).
The SA was run for four sample sizes $k\in\{4, 8, 16, 32\}$ which are referred to as SA$_k$.

The benefit of the surrogate model is supported by a comparison with two more methods denoted the WIRIS-EA and IRIS-EA.
The WIRIS-EA is the WISM-EA (including the caching) with a sole exception of using IRIS instead of its neural network surrogate model to compute the partial costs $\overline{C}(\Sigma_{w,i})$.
The IRIS-EA simply combines the evolutionary algorithm presented in Section~\ref{sec:method_ea} with IRIS approximation of the closed DTP of the complete sequence $\Sigma$, i.e., $C^*(\Sigma)$.

\revise{}{Due to computational requirements of IRIS, its sampling precision was reduced to $k_\text{max} = 16$ for both WIRIS-EA and IRIS-EA.
However, in many cases, we were unable to get results competitive to other methods for lower values of the $\stopTime$.
Nevertheless, the value of $k_\text{max}$ was selected empirically based on the overall best tradeoff between the DTP length approximation precision and the number of evaluations achievable in $\stopTime$.
Notice that the final IRIS call (Line~\ref{alg:ea_dtp}, Algorithm~\ref{alg:ea}) was still executed using $k_\text{max} = 1024$ to show impact of the found sequences by the EA.

Both the methods significantly underperform the WiSM-EA with the windowing WIRIS-EA being a better alternative.
The reason is evident from aggregated results over all densities for $n=100$, where the WiSM-EA achieves a significantly higher rate of $75.4\times 10^3$ evaluations of the DTP solution cost per second in comparison to much slower WIRIS-EA and IRIS-EA with $221$ and $7.84$ evaluations per second, respectively.
{}

Examples of the found solutions are shown in Fig.~\ref{fig:solution_examples} and a detailed report of the evaluation follows.

\begin{figure}[t]
   \includegraphics[width=\columnwidth, valign=c]{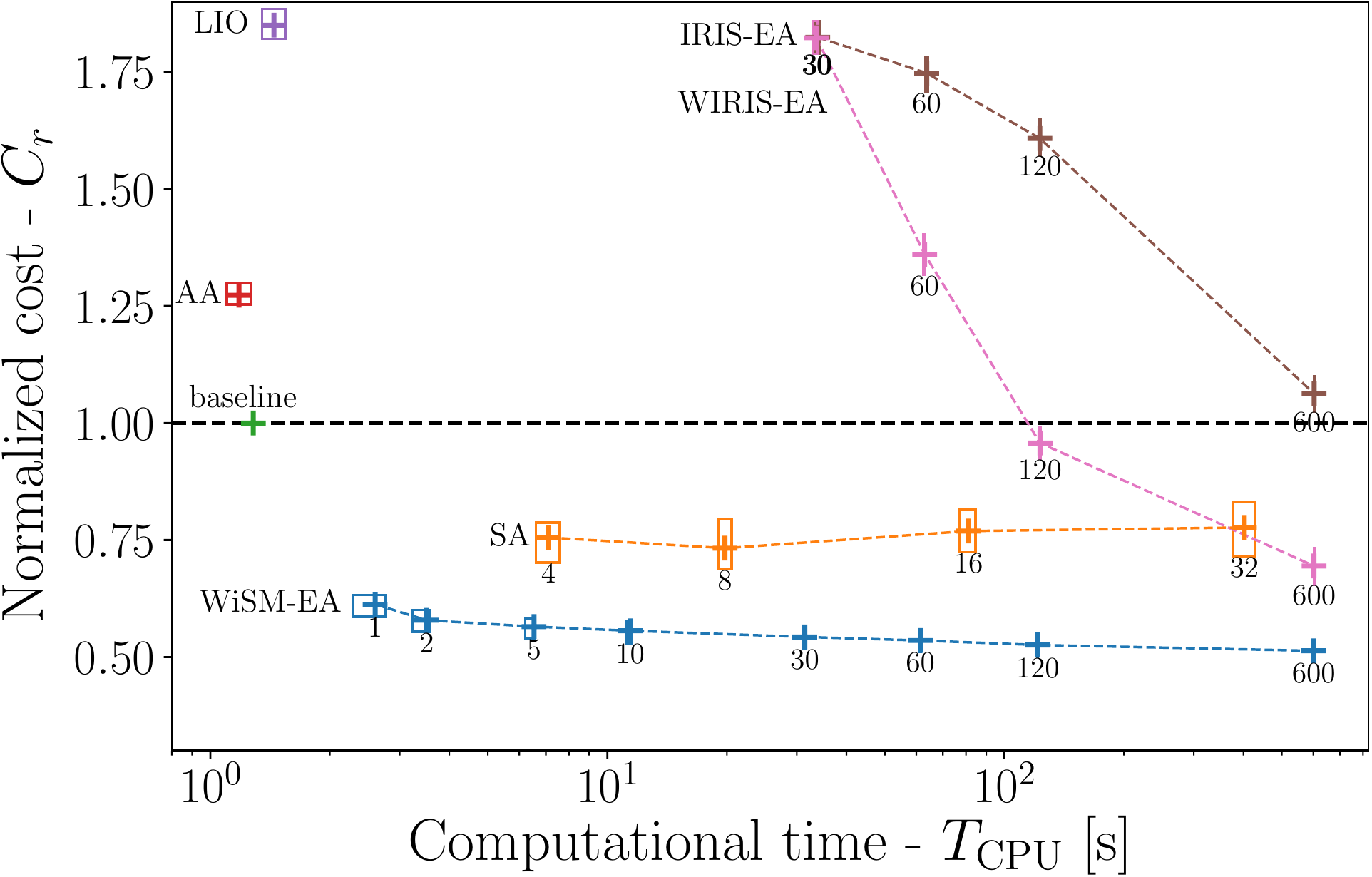}
   \caption{
      Normalized cost $C_r$ and real computational requirements for DTSP instances with the density $d=1.0$ and $n=100$ target location.
      Several values for the WiSM-EA indicate different time limit and for SA the particular number of the samples.
      Note the logarithmic scale of the time axis.
      \label{fig:results_d1_n100}
   }
\end{figure}

{
   \begin{figure*}[!htb]
      \subfloat[high density instances, $d=10.0$, $n=100$\label{fig:results_high}]{\includegraphics[width=\columnwidth]{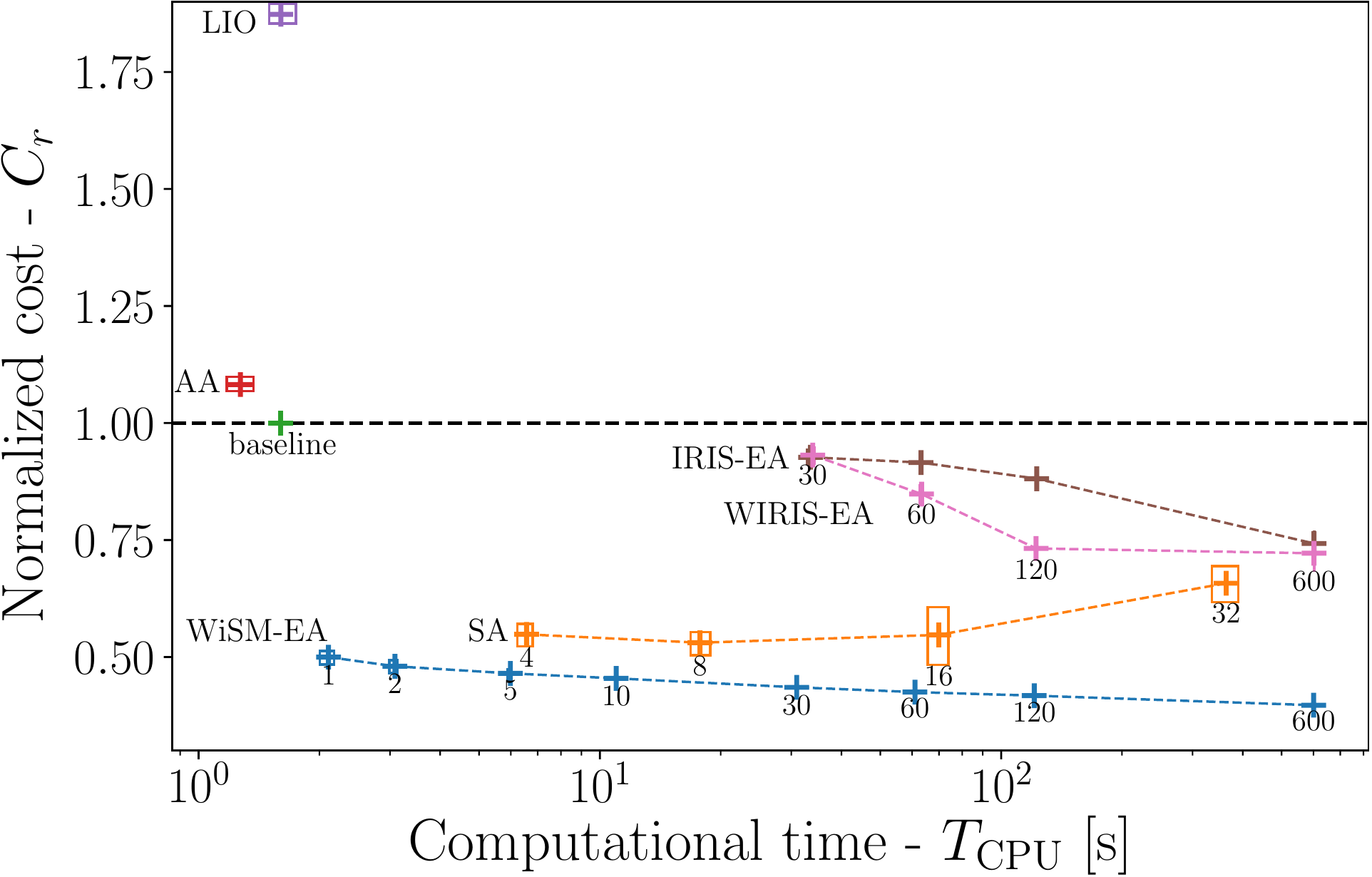}} 
      \hfill
      \subfloat[low density instances, $d=0.1$, $n=100$\label{fig:results_low}]{\includegraphics[width=\columnwidth]{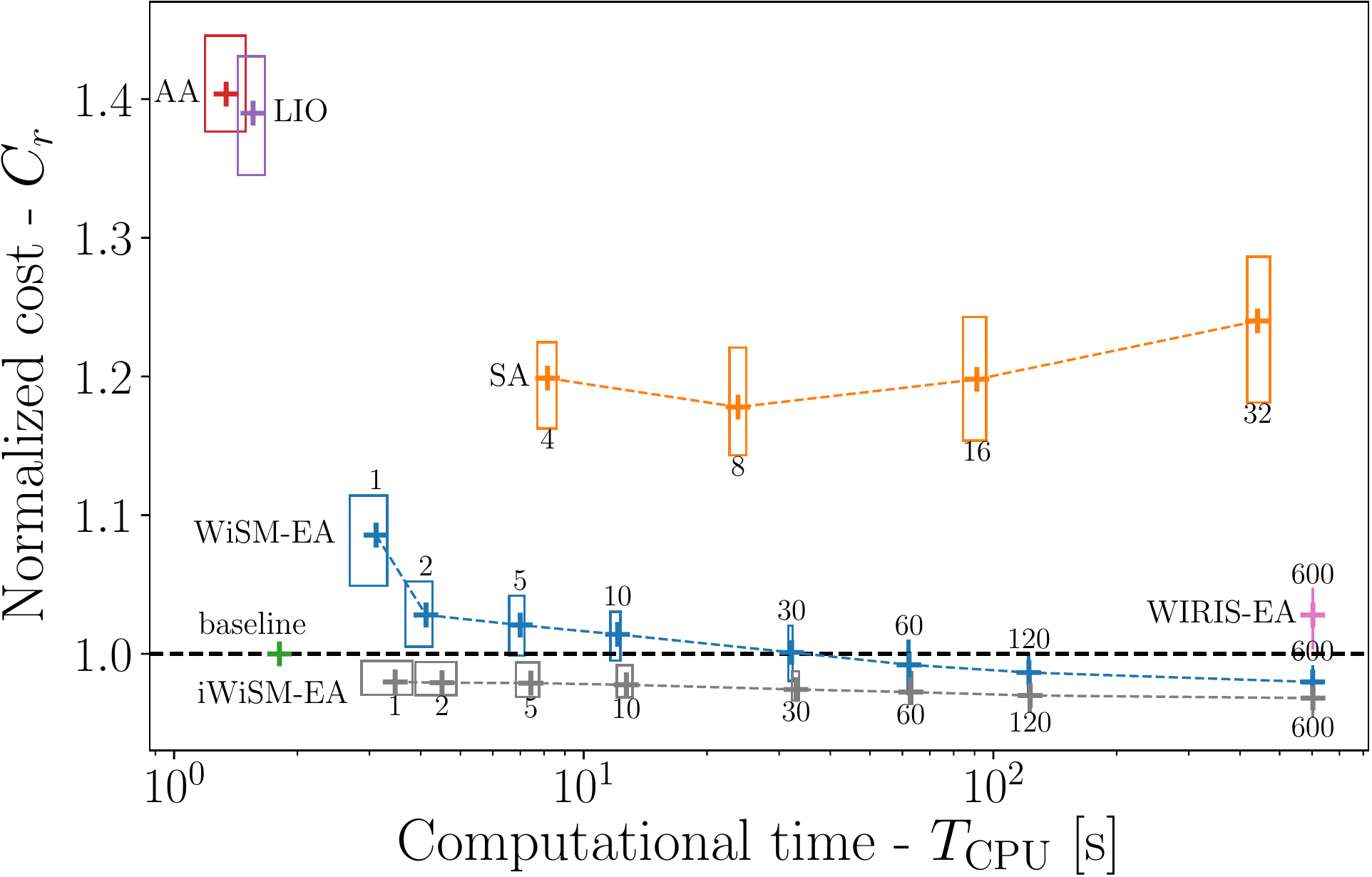}} 
      \caption{
     Performance of the evaluated DTSP solvers in high ($d=10$) and low ($d=0.1$) density instances with $n=100$ target locations.
     The achieved normalized cost $C_r$ is shown according to the required computational time or a given time limit (for the WiSM-EA).
     For $d=0.1$, the iWiSM-EA refers to the proposed WiSM-EA with the initialization of the population by the solution of the Euclidean TSP provided by the Concorde solver.
     \label{fig:results_densities}
      }
   \end{figure*}
}

{
   \begin{figure*}[!htb]
      \subfloat[small instances $n=25$, $d=1.0$]{\includegraphics[width=\columnwidth]{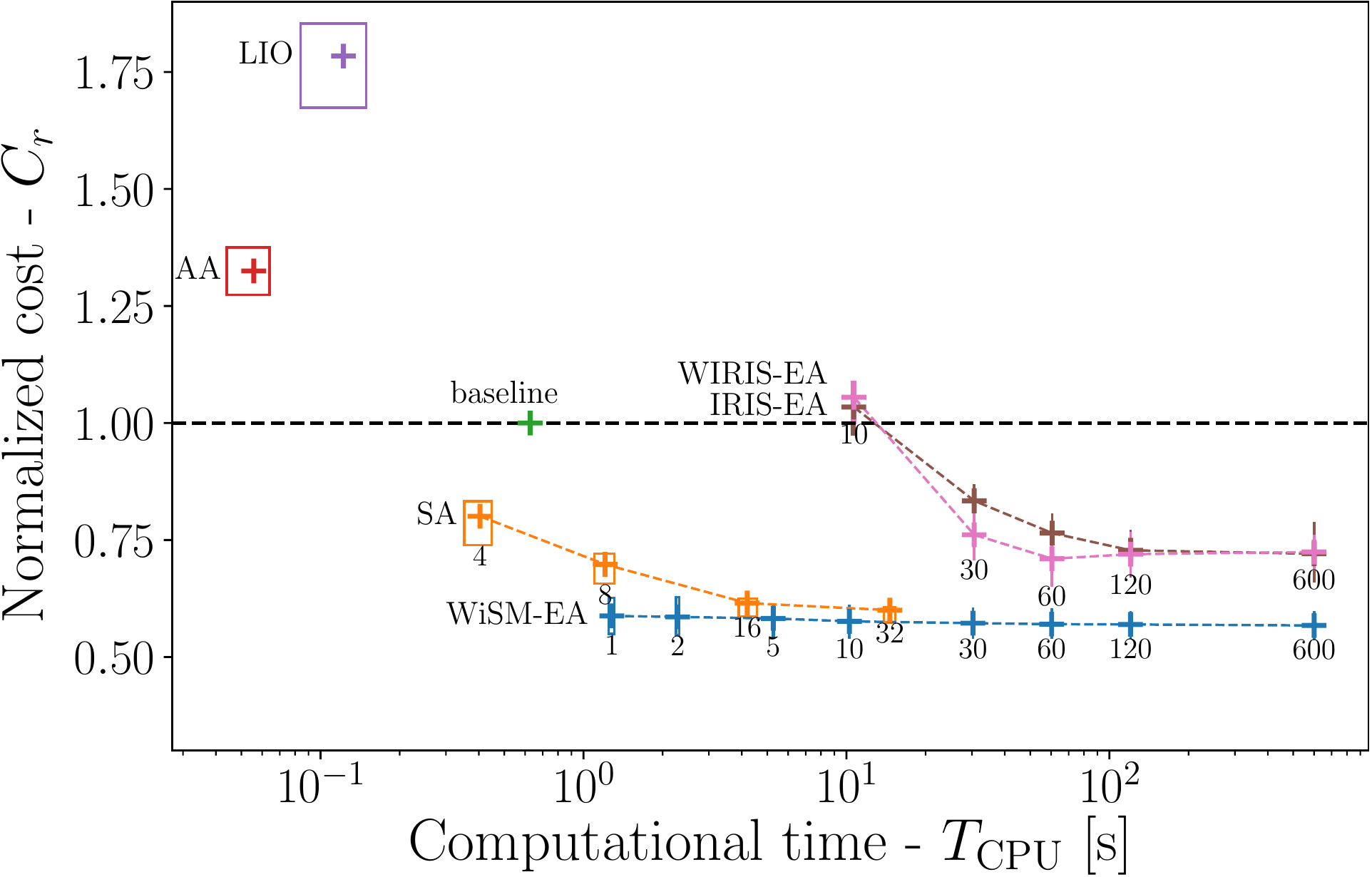}}
      \hfill
      \subfloat[large instances $n=500$, $d=1.0$]{\includegraphics[width=\columnwidth]{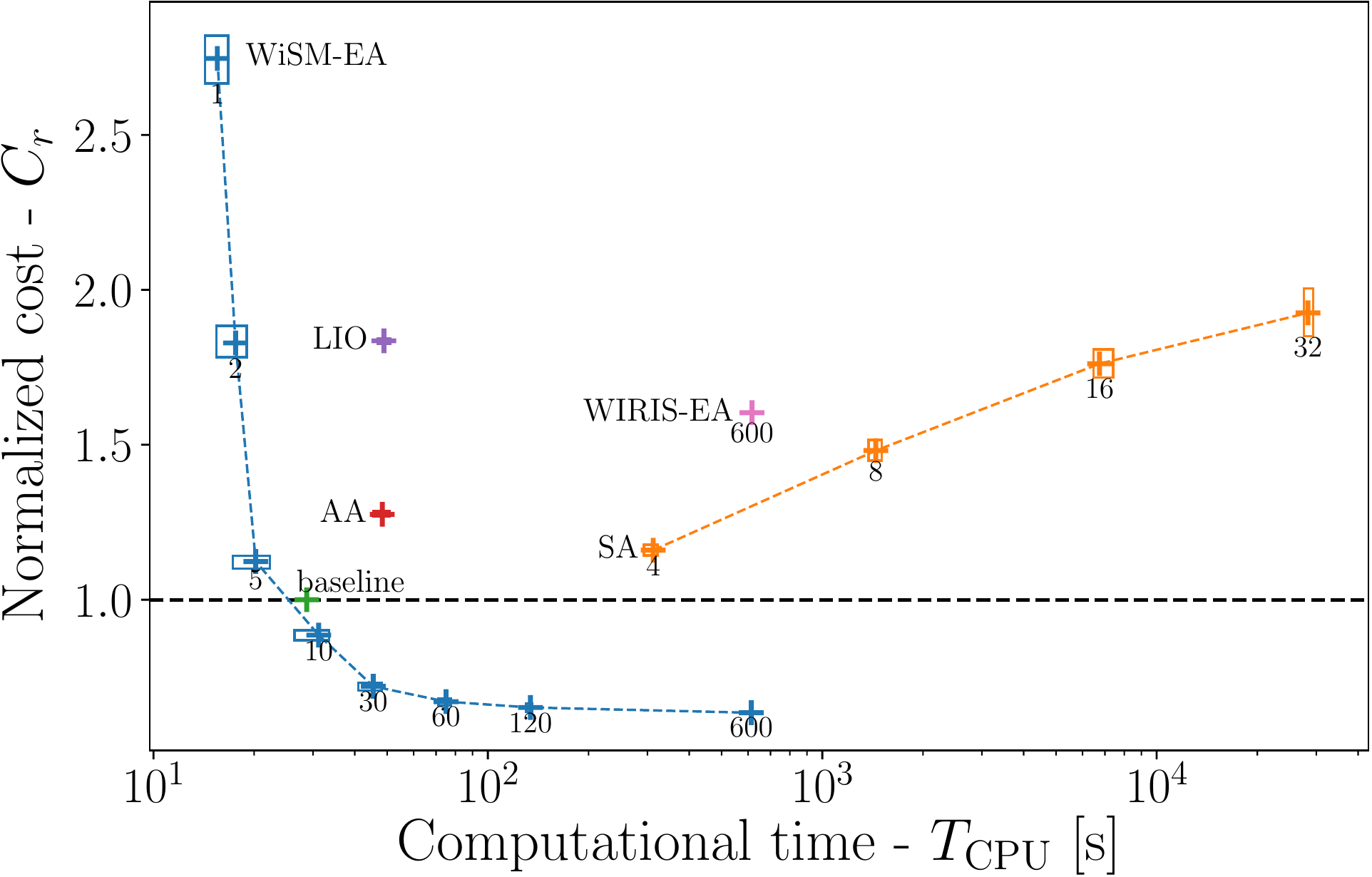}}
      \caption{
     Performance of the evaluated DTSP solvers in small ($n=25$) and large ($n=500$) instances with the density $d=1.0$.
     \label{fig:results_locations}
      }
   \end{figure*}
}

{
\begin{figure*}[t]
   \subfloat[WiSM-EA, $d=10$, $C=125.09$]{\includegraphics[width=0.3\textwidth,trim={0.6cm 0.6cm 0 1cm},clip]{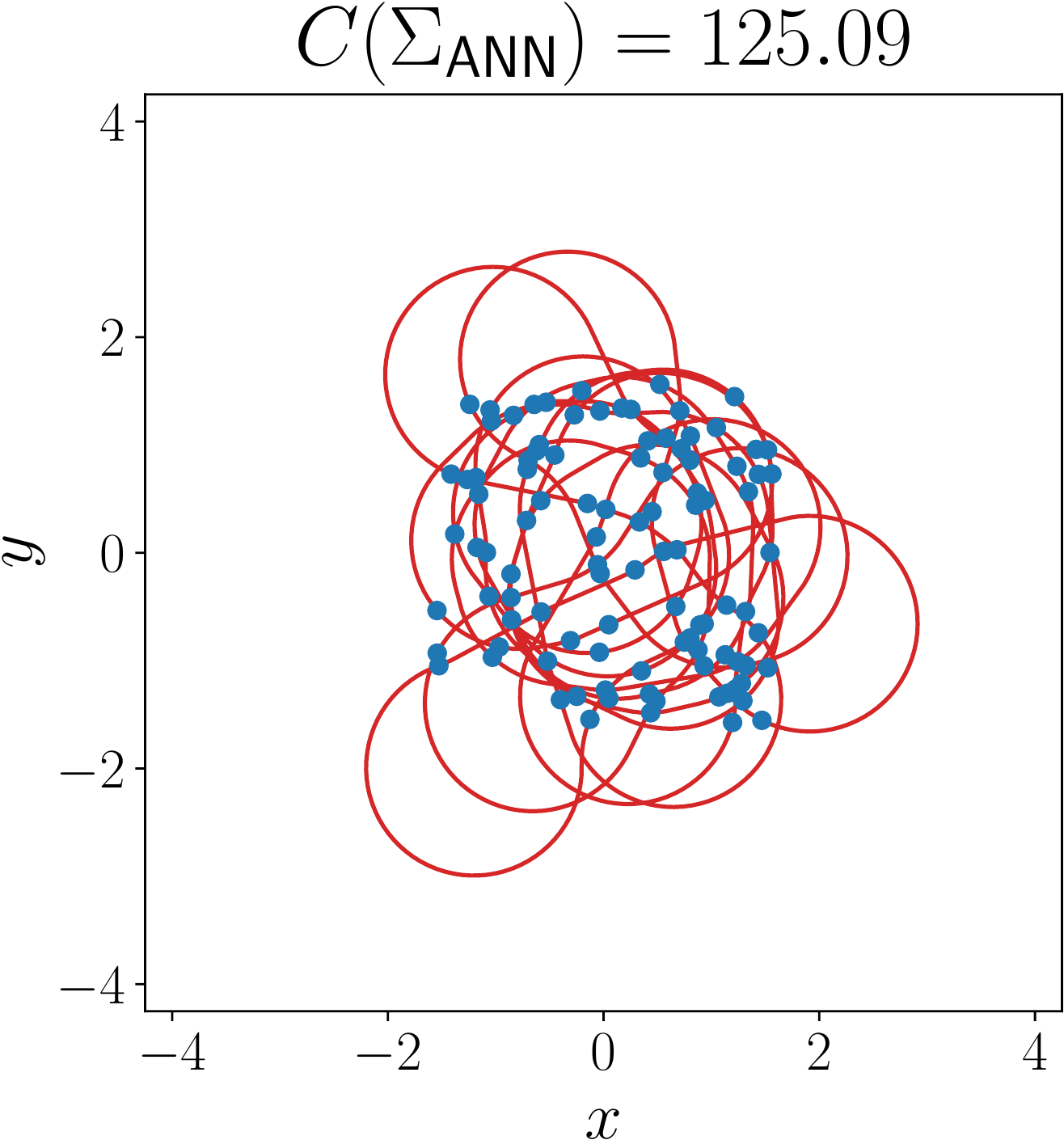}}
   \hfill
   \subfloat[WiSM-EA, $d=1$, $C=147.37$]{\includegraphics[width=0.3\textwidth,trim={0.6cm 0.6cm 0 1cm},clip]{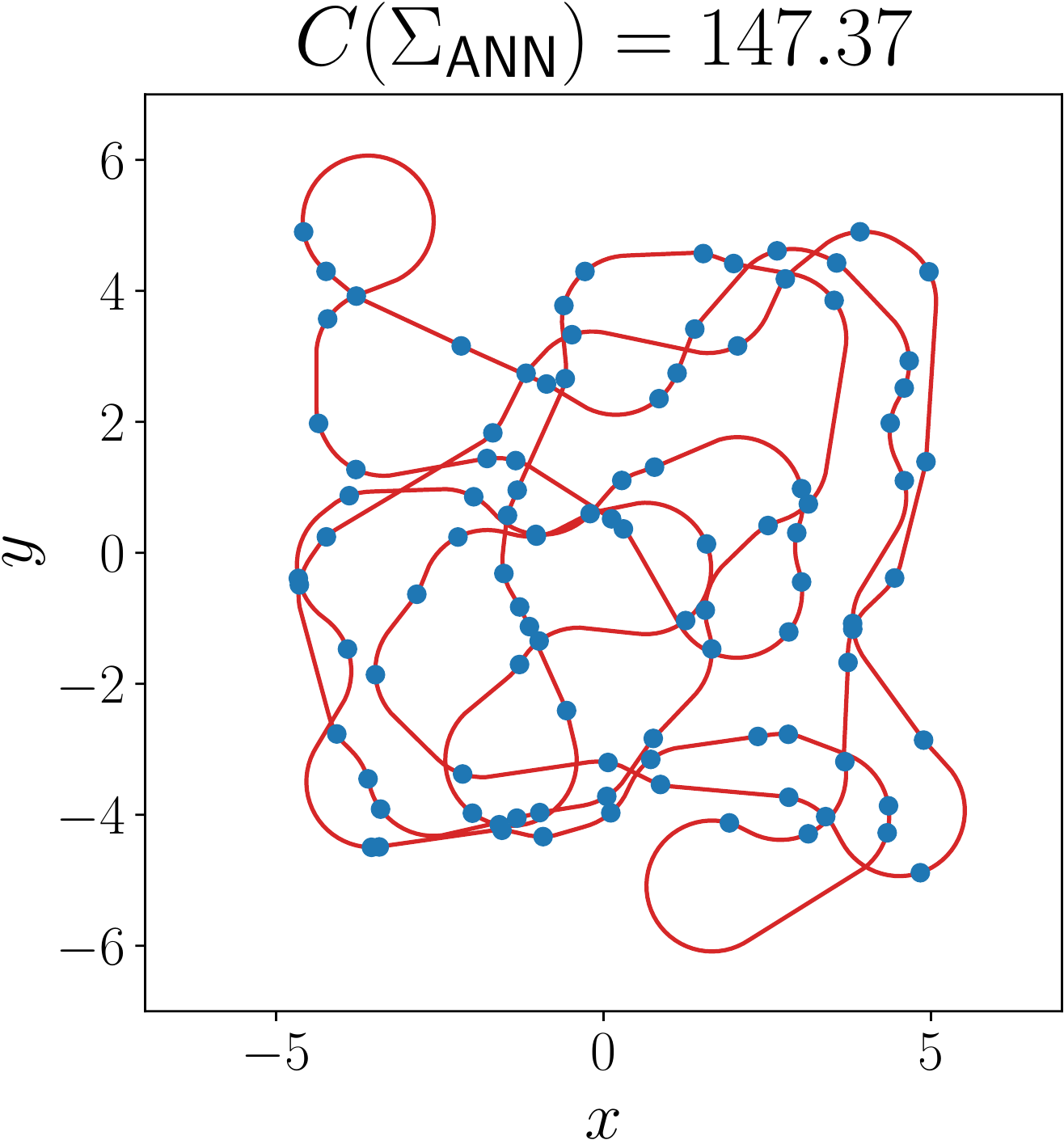}}
   \hfill
   \subfloat[WiSM-EA, $d=0.1$, $C=283.73$]{\includegraphics[width=0.3\textwidth,trim={0.6cm 0.6cm 0 1cm},clip]{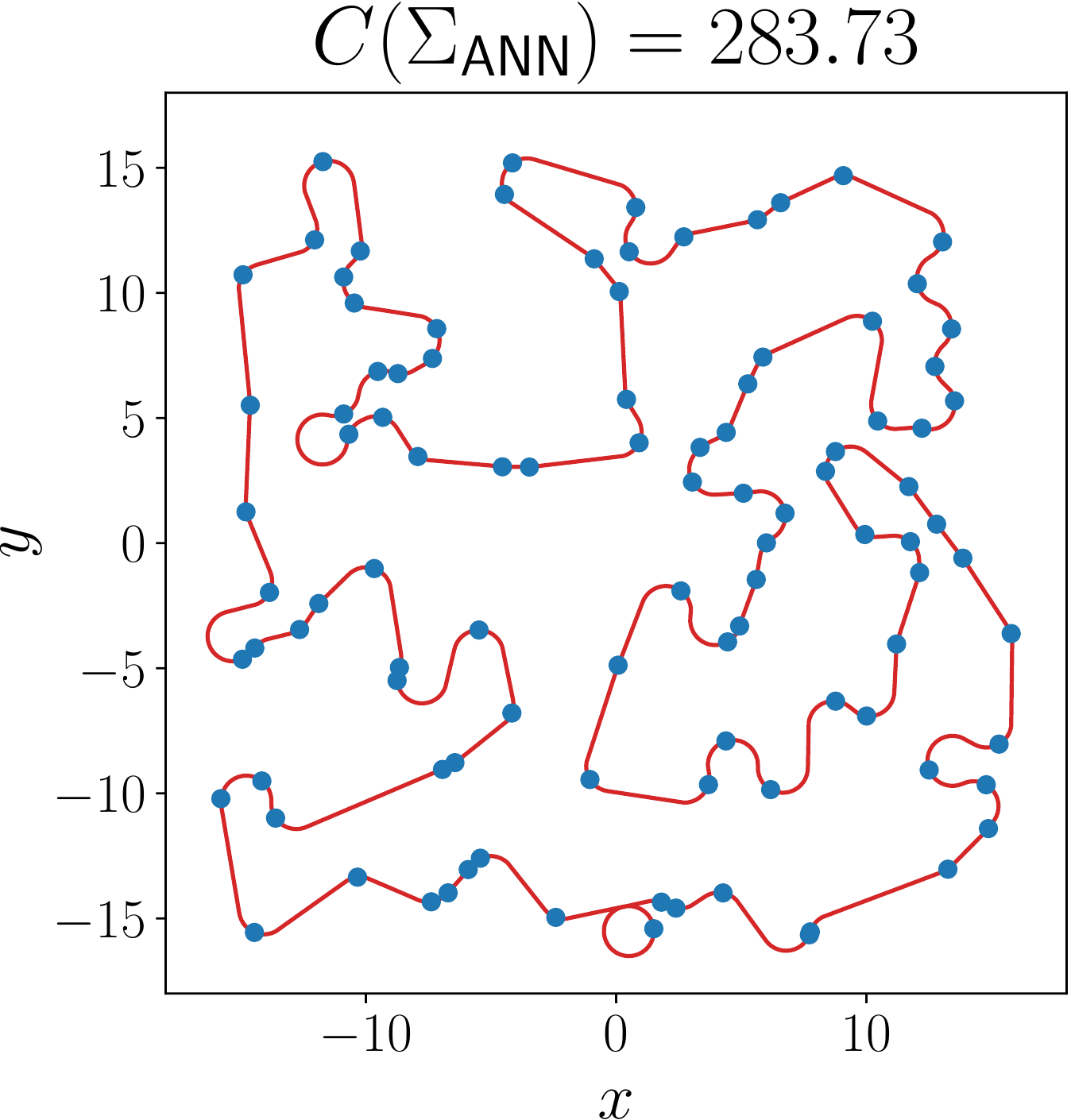}}

   \subfloat[Baseline, $d=10$, $C=321.32$]{\includegraphics[width=0.3\textwidth,trim={0.6cm 0.6cm 0 1cm},clip]{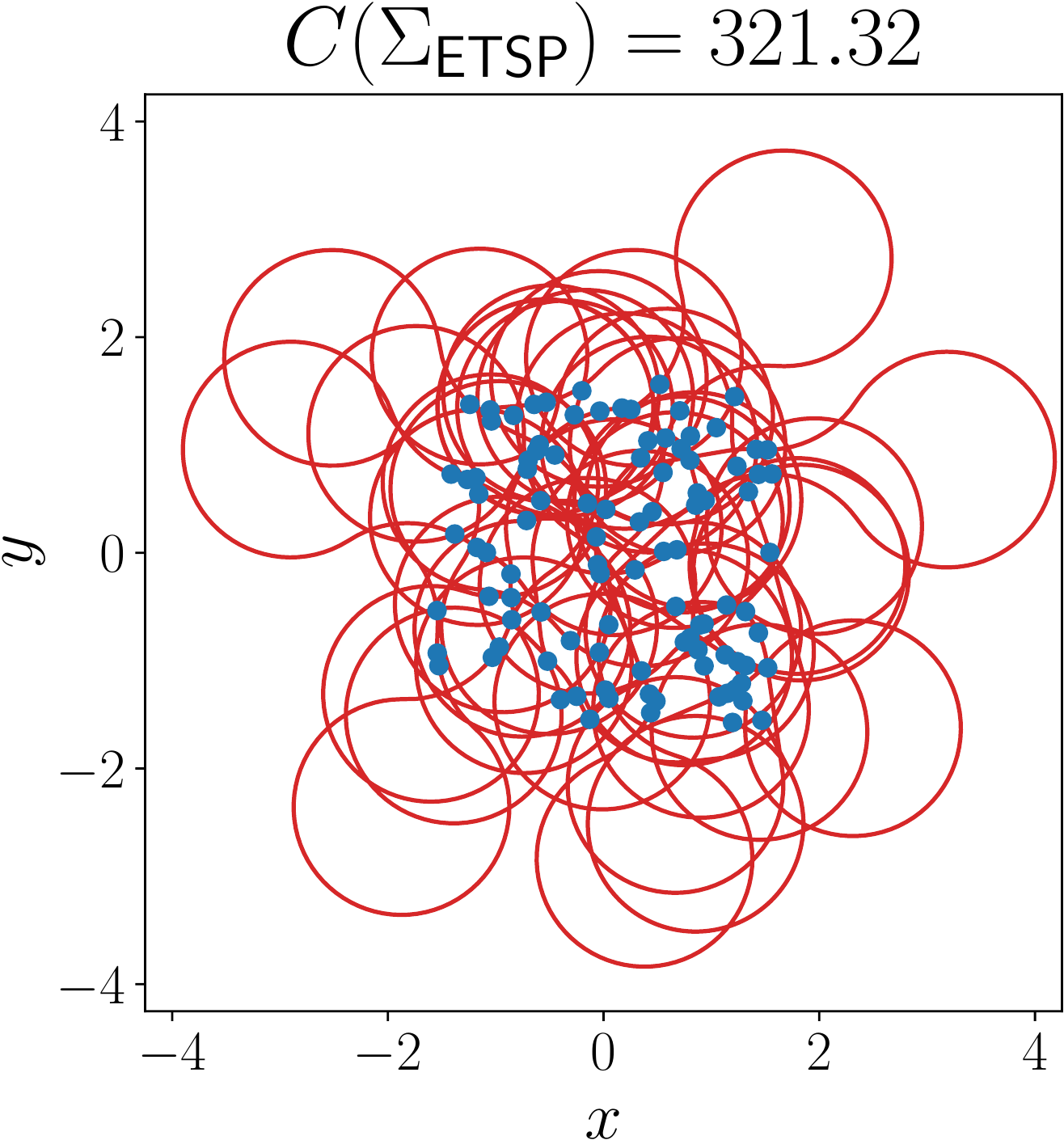}}
   \hfill
   \subfloat[Baseline, $d=1$, $C=280.28$]{\includegraphics[width=0.3\textwidth,trim={0.6cm 0.6cm 0 1cm},clip]{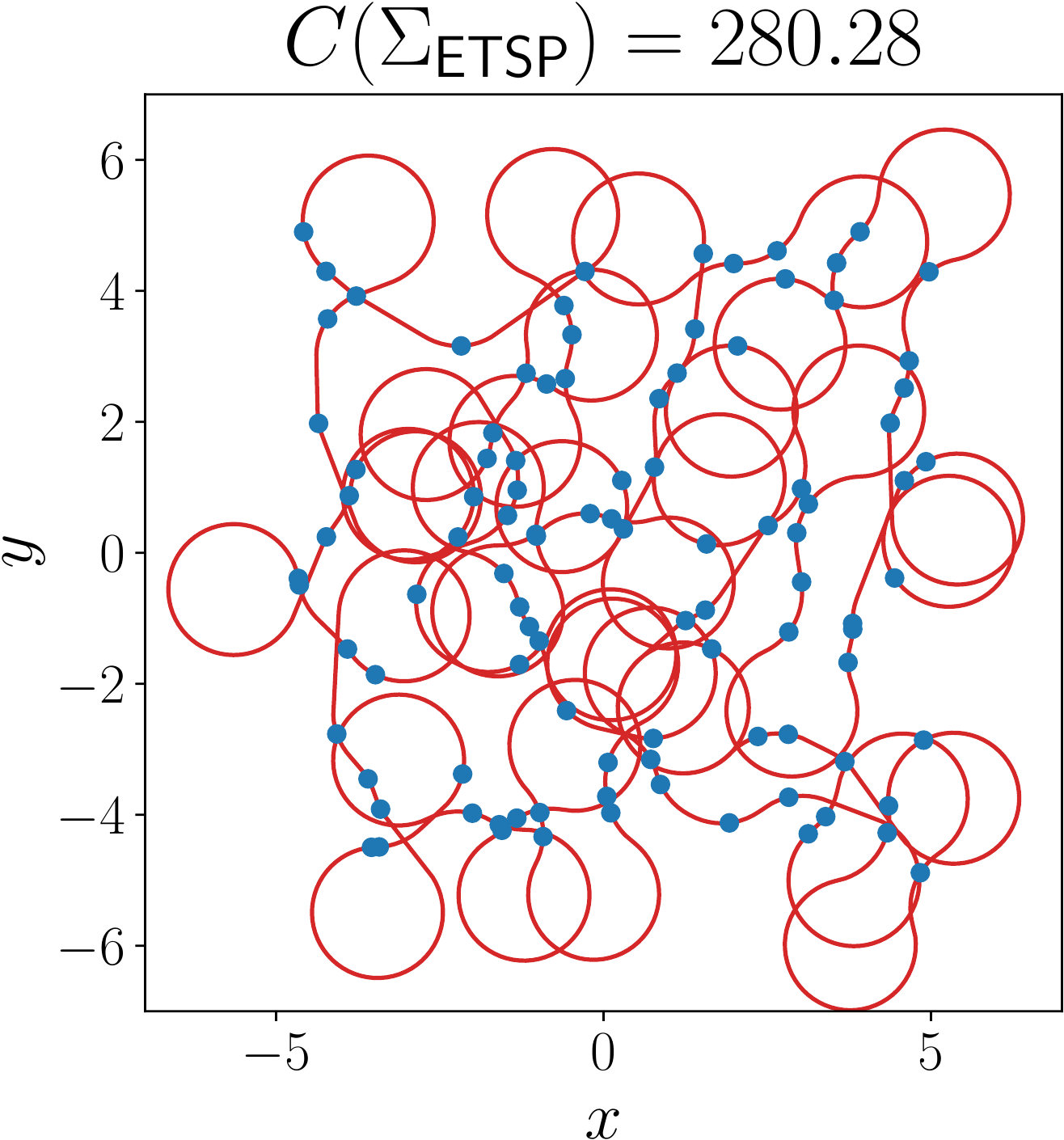}}
   \hfill
   \subfloat[Baseline, $d=0.1$, $C=289.35$]{\includegraphics[width=0.3\textwidth,trim={0.6cm 0.6cm 0 1cm},clip]{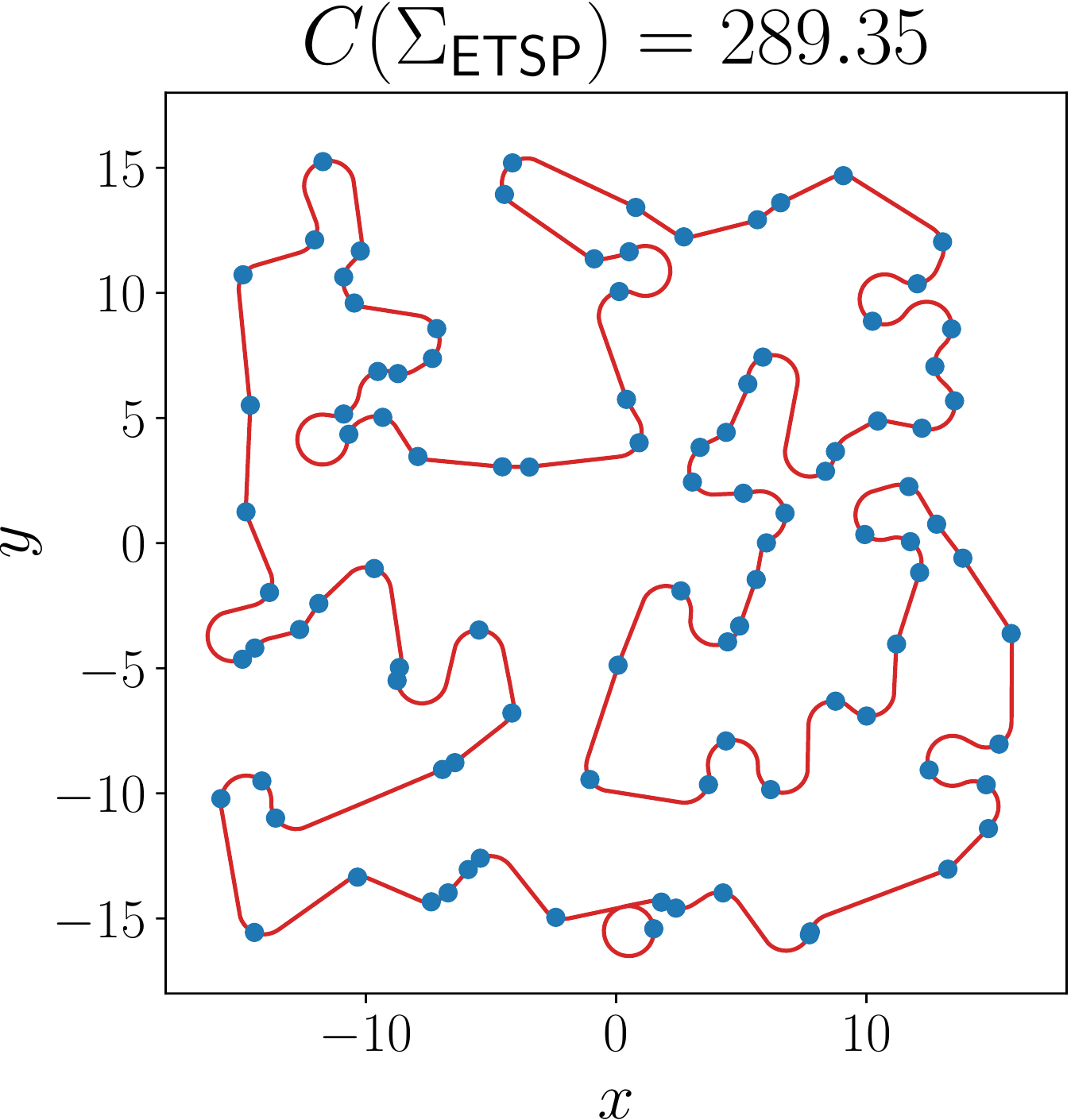}}
   \caption{
      Solutions of the DTSP with $n=100$ target locations and density $d$ found by the proposed WiSM-EA (upper) and baseline (bottom).
      The target locations are depicted as small blue disks, and the found Dubins tour is in the red.
      Each column represents the same single instance, and thus the solutions can be directly compared.
      The final cost $C$ is computed using the IRIS with the maximum resolution $k_\text{max}=1024$.
      \label{fig:solution_examples}
   }
\end{figure*}
}

The achieved results for the DTSP instances with the density $d=1$ and $n=100$ locations are shown in Fig.~\ref{fig:results_d1_n100}, where the Pareto front is formed by three methods only: the AA, baseline, and the proposed WiSM-EA.
{Taking only the shortest $\stopTime=\SI{1}{s}$ limit into consideration, all three methods give results in units of seconds.
In particular, the AA in $\Tcpu=\SI{1.2}{s}$, the baseline in $\Tcpu=\SI{1.3}{s}$, and the $\text{WiSM-EA}$ in $\Tcpu=\SI{2.6}{s}$.}
The WiSM-EA achieves a significantly better $C_r=0.61$ than the AA with $C_r=1.3$ and the baseline with $C_r=1.0$.
Besides, the normalized cost further improves to $C_r=0.51$ in approximately ten minutes (WiSM-EA\textsubscript{600}).
The only rival in the solution cost is the SA method with the best value $C_r=0.73$ achieved by the SA\textsubscript{8} in $\Tcpu=\SI{20}{s}$. 
The solution cost provided by the SA should theoretically decrease with the increasing number of samples; however, the observed behavior is different, and the opposite trend can be noticed for the SA\textsubscript{16} and SA\textsubscript{32}.
This behavior can be explained by too large instances of the underlying Generalized TSP and heuristic solution of the transformed TSP instance. 
Similar behavior was observed for all tested instances having $n \geq 100$.

Evaluation results for the DTSP instances with $n=100$ locations and high and low densities are depicted in Fig.~\ref{fig:results_densities}.
For the high density instances with $d=10$, the results are similar to $d =1$ (compare Fig.~\ref{fig:results_d1_n100} and Fig.~\ref{fig:results_high}).
The AA reaches the mean $C_r=1.08$, while the proposed WiSM-EA achieves significantly better results ranging from $C_r=0.5$ (WiSM-EA\textsubscript{1}) to $C_r=0.4$ (WiSM-EA\textsubscript{600}).
LIO does not perform well, while the SA provides competitive results for the time limit around \SI{10}{s}.

For the low-density instances with $d=0.1$, the DTSP becomes closer to the corresponding Euclidean TSP because locations are relatively far from each other and the minimum turning radius constraint does not significantly affect the length of the final Dubins tour.
It is also indicated by the results for the baseline with the sequence determined by the ETSP solver.
The proposed WiSM-EA outperforms all other solvers, but the baseline is outperformed only for the time limits $\stopTime\ge \SI{30}{s}$.
Based on that, we employed a solution of the corresponding Euclidean TSP provided by the Concorde solver~\cite{concorde} in the initialization of the whole initial generation in the \initializePopulation method (Line~\ref{alg:ea_init}, Algorithm~\ref{alg:ea}) of the proposed EA instead of a random permutation to improve the performance.
The modified algorithm is denoted iWiSM-EA in Fig.~\ref{fig:results_low}.
Although the iWiSM-EA provides outstanding performance in low-density instances, it performs similarly to the WiSM-EA for $d=1$ and $d=10$. 
For large instances, we observed slightly higher final costs, which can be explained by the zero diversity of the initial population making the optimization prone to local minima (data not shown for readability).
We conclude that the initialization of the iWiSM-EA approach is beneficial for low-density instances only.

Computational performance for instances with small ($n=25$) and high ($n=500$) number of locations is shown in Fig.~\ref{fig:results_locations}.
For small instances, the Pareto front is formed by the AA, SA, and WiSM-EA solvers. 
Note that WIRIS-EA achieves the best result for $\stopTime = \SI{30}{s}$ while its performance deteriorates for longer time limits which can be explained by the reduced precision using $k_\text{max} = 16$.
Solutions found by the WiSM-EA\textsubscript{1} have the mean cost $C_r=0.59$ that is significantly lower than for the AA with $C_r=1.32$, the SA\textsubscript{4} with $C_r=0.8$, but also the SA\textsubscript{8} with $C_r=0.7$ with the mean computational time $\Tcpu=\SI{1.3}{s}$.
The AA is the fastest approach with the mean $\Tcpu=\text{\SI{0.06}{s}}$ and the SA\textsubscript{4} needs $\Tcpu=\text{\SI{0.4}{s}}$.
For large instances with $n=500$, the proposed WiSM-EA dominates the other methods.
The WiSM-EA\textsubscript{10} provides the mean $C_r=0.89$ in $\Tcpu=\SI{21}{s}$ and reaches $C_r=0.64$ in approximately ten minutes.

\subsection{Discussion and Possible Future Work\label{sec:sec:experiments_discussion}}
Based on the reported results, the proposed approximator WiSM is a vital approach that enables the solution of the DTSP instances by a relatively simple evolutionary algorithm.
The developed WiSM-EA scales significantly better than the other evaluated methods in both the problem size $n$ and density $d$. 
For large instances, it dominates the other methods in the computational requirements and the quality of the found solutions.
From a practical point of view, the WiSM-EA gives the best results in units of seconds for small and medium-sized instances, while for the large instances with hundreds of locations, it needs only low tens of seconds.
To improve the rate of convergence for low-density instances of the DTSP, we suggest using the ETSP initialization of the WiSM-EA.

Regarding the other methods, the AA is the fastest algorithm, but it always provides $C_r>1$ and thus worse solutions than the baseline.
The SA provides competitive results to the proposed WiSM-EA in medium and high-density instances, but it does not scale with the problem size and the number of the utilized samples, mostly because of the limitations of the underlying solver to the transformed Generalized TSP. 

Although it is not the aim of this paper, our initial experiments indicated (data not shown), that WiSM is robust w.r.t. different neural network architectures. 
Nevertheless, future research should focus on finding possibly more effective regressors. 
The similar applies to approaches generating representative training data for WiSMs learning. 

Regarding the future deployments of the proposed surrogate approximator of the Dubins tour costs, we believe it can also be utilized in other Dubins routing problems such as the Dubins Orienteering Problem~\cite{penicka17ral} where many sequences have to be evaluated.
Moreover, the model can be generalized for touring problems with neighborhood areas instead of single locations, where the recently introduced Generalized Dubins Interval Problem~\cite{vana2018gdip} can be utilized for the model learning using high-quality solutions, and then for solving the Dubins TSP with Neighborhoods~\cite{faigl19jfr}.
Finally, the herein presented results motivate for future work on the approximation of costs of curvature-constrained tours for more complex vehicle models than Dubins vehicle, e.g., Bézier curves~\cite{faigl18ral}.


\section{Conclusion\label{sec:conclusion}}

We present a novel approach to solve the Dubins Traveling Salesman Problem (DTSP) by a relatively simple Evolutionary Algorithm that is based on the proposed surrogate approximator of the Dubins tour cost called WiSM.
Even though collecting enough training data might take considerable time, once the dataset is built and the WiSM is trained, it can provide Dubins tour cost estimates in a very fast rate, which can be exploited by robust global search methods such as Evolutionary Algorithms. 
The developed WiSM-EA has been evaluated on DTSP instances of varying size and also the density of the locations to be visited.
Based on the reported results, the WiSM-EA outperforms the existing state-of-the-art approaches in the quality of the found solutions and also computational requirements.
The results demonstrate the proposed method scales with the problem size and density, and thus, it is a suitable heuristic for finding high-quality solutions of curvature-constrained routing problems with Dubins vehicle.

\bibliographystyle{IEEEtran}
\bibliography{main}
\end{document}